# Research on Early Warning Model of Cardiovascular Disease Based on Computer Deep Learning

Yuxiang Hu[1], Jinxin Hu[2], Ting Xu[3], Bo Zhang[4], Jiajie Yuan[5], Haozhang Deng[6]
[1]Johns Hopkins University, USA, yhu102@jhu.edu
[2]Arizona State University, USA, jinxinhu@asu.edu
[3]University of Massachusetts Boston, USA, ting.xu001@umb.edu
[4]Texas Tech University, USA, zhangbo5782@gmail.com
[5]Brandeis University, USA, jiajieyuan@brandeis.edu
[6]Northwestern University, USA, haozhangdeng2021@u.northwestern.edu

*Abstract*—This project intends to study a cardiovascular disease risk early warning model based on one-dimensional convolutional neural networks. First, the missing values of 13 physiological and symptom indicators such as patient age, blood glucose, cholesterol, and chest pain were filled and Z-score was standardized. The convolutional neural network is converted into a 2D matrix, the convolution function of 1,3, and 5 is used for the first-order convolution operation, and the Max Pooling algorithm is adopted for dimension reduction. Set the learning rate and output rate. It is optimized by the Adam algorithm. The result of classification is output by a soft classifier. This study was conducted based on Statlog in the UCI database and heart disease database respectively. The empirical data indicate that the forecasting precision of this technique has been enhanced by 11.2%, relative to conventional approaches, while there is a significant improvement in the logarithmic curve fitting. The efficacy and applicability of the novel approach are corroborated through the examination employing a one-dimensional convolutional neural network.

## I. Introduction

Heart disorders represent a prevalent category of long-term illnesses both in urban and rural settings across the USA. Consequently, the implementation of precise early alert systems carries immense practical importance for the mitigation and management of such conditions. Although angiography can make the correct diagnosis of cardiovascular disease in the clinic, it is not only expensive but also causes damage to the human body [1]. In addition, ECG and related integral indicators are commonly used in clinical prediction, but this evaluation has a strong theoretical basis and practical experience for physicians [2]. Recently, some scholars have suggested using machine learning and other methods to model risk factors in patients' medical records[3]. This kind of treatment is not only harmless, but also economical, and has gradually become a research topic of great practical value and challenge.

Electronic medical records contain a large number of patient information, including patient diagnosis, doctor's orders, physical examination reports, vital signs, and demographics [4]. The existing patient image representation technology has high requirements, is limited by the personal experience of researchers and specific electronic medical record systems, and has some problems such as scalability and low universality [5]. In recent years, in the research related to image feature extraction, some scholars have applied it to the image feature representation, such as the unique hot zone and word package. However, this sparse representation-based approach often fails to reflect the semantic information and sequence information in medical records [6]. However, for time series learning, CNN can only capture part of the features, and it must be assumed that all the data in the medical record follows the time series [7]. The algorithm of periodic neural network, such as long-term memory neural network, adopts different "thresholds" to extract data, which has high efficiency for electronic medical records with serial characteristics [8]. Some studies have used the representation learning algorithm based on a bidirectional loop neural network to improve the expression learning performance and interpretability of the model by introducing attention mechanism and other means [9]. Although this new research approach can improve the probability of disease occurrence, it does not take into account the differences between different cases [10]. Some studies have shown that independent models are used for representation learning of various types of data to improve the recognition accuracy [11]. However, in fact, each treatment has its own corresponding one or more conditions [12]. Although this method can be a good risk prediction, it will be reduced in some aspects because it does not take into account the correlation between the condition and drug use.

This project intends to construct a new method combining recurrent neural network and attention mechanism to establish a new risk prediction method for cardiovascular diseases. RPMC can automatically extract high-quality feature representations from massive, heterogeneous, time-series electronic medical records for precise cardiovascular disease risk assessment [13]. This project proposes a modeling method that employs a novel graph convolutional neural network application, verified to enhance the model's learning performance on time-series data and to open up new pathways for accurately predicting patient survival rates [14]. In addition, aiming at the difference and correlation of multi-source information, this project proposes a representation method based on multi-source information [15]. Finally, the risk factors of cardiovascular system can be predicted by integrating the characteristic information of multiple modules.

## II. ARCHITECTURE OF ONE-DIMENSIONAL CONVOLUTIONAL NEURAL NETWORKS

Different from other neural networks, CNNS can reduce parameters in the network through local connectivity and weight sharing, and be more sensitive to changes in details, so as to better obtain the required features [16]. People construct a new deep convolutional neural network VGGNet, optimize it accordingly, and finally form a complete set of deep learning models. First, the CNN model takes the input matrix of each pixel in the image as input, and processes the image [17]. The images are in black and white: one in gray and the other in color [18]. The grayscale image is described by a two-dimensional matrix, because the color of each pixel has the same value, so it is also called "single channel"; Color maps, also known as RGB maps, are represented by a three-dimensional matrix, with each dimension representing red, green, and blue [19]. An overall architecture of a CNN-based cardiovascular risk early warning system is presented (Figure 1 cited in Sensors 2023, 23(14), 6507). The method takes 2D eigenmatrix as input, and adopts a method similar to the gray level, each sample is a single channel.

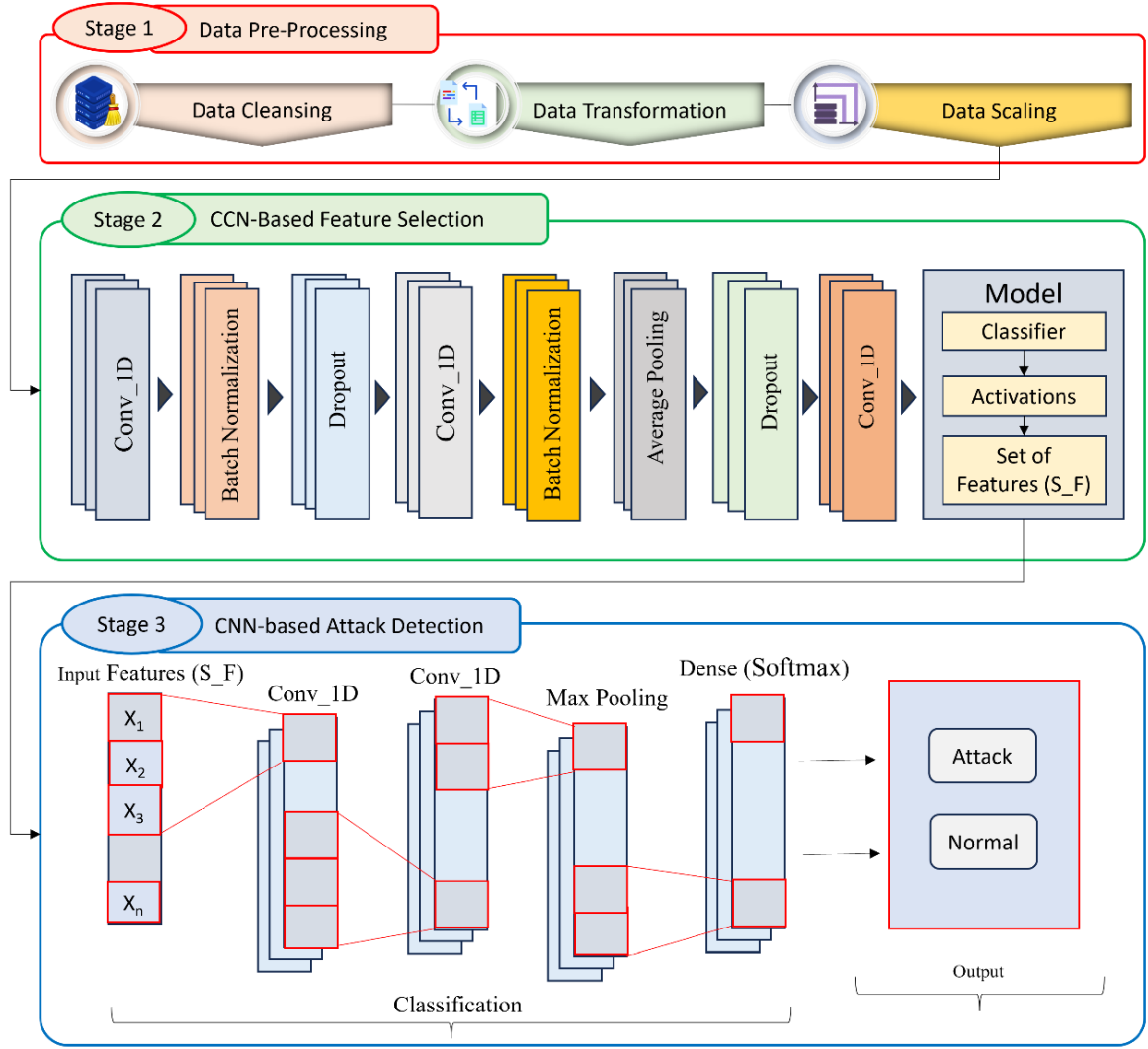

Fig. 1. 1D-CNN overall framework structure

## A. Input layer design

The input stage completes the training of 13 parameters, and the 13-parameter property matrix is imported into the convolutional network and connected to the following convolutional network [20]. Because the data selected in this experiment is 13 dimensions, the parameter vectors of the 13 parameters are expressed by $F_1, F_2, L, F_{13}$, and these parameters are sorted from top to bottom in the order of a column vector, thus obtaining the characteristic matrix representing the data.

$$F = \begin{pmatrix} F_1 \\ F_2 \\ M \\ F_{13} \end{pmatrix} = \begin{pmatrix} f_{11} & f_{12} & & f_{1y} \\ f_{21} & f_{22} & L & f_{2y} \\ M & M & & M \\ f_{131} & f_{132} & L & f_{13y} \end{pmatrix} \quad (1)$$

## B. Convolutional layer design

Convolution kernel is a key link in convolutional neural network, which convolves the eigenmatrix of input layer through convolution function to obtain deep information [21]. The convolution kernel can be viewed as a weighted matrix of $x \times y$, which can be expressed as:

$$\eta = \begin{bmatrix} f_{11} & f_{12} & & f_{1y} \\ f_{21} & f_{22} & L & f_{2y} \\ M & M & & M \\ f_{x1} & f_{x2} & L & f_{xy} \end{bmatrix} \quad (2)$$

Because the convolution core used here is one-dimensional, people set the height $x$ to 1 and the width $y$ to 1,3,5. The convolutional kernel is convolved from left to right in the order of length 1, and new properties are obtained after each convolution [22]. If the parameters that constitute each parameter are expressed by $G_i$, that is, the parameter vector of the first parameter value in the convolution window, then the eigenmatrix in the window can be expressed as:

$$G_i = \begin{pmatrix} G_i \\ G_{i+1} \\ M \\ G_{i+y-1} \end{pmatrix} = \begin{bmatrix} g_{i1} & g_{i2} & & g_{iy} \\ g_{(i+1)1} & g_{(i+1)2} & L & g_{(i+1)y} \\ M & M & & M \\ g_{(i+y-1)1} & g_{(i+y-1)2} & L & g_{(i+k-1)1=y} \end{bmatrix} \quad (3)$$

The eigenvalues of each step can be obtained from formula (4):

$$C_i = g(\eta * G_i + \varepsilon) = g\left(\sum_{d=1}^{x}\sum_{s=1}^{y} \mu_{ds} g_{(i+d-1)s} + \varepsilon\right) \quad (4)$$

$\varepsilon$ is the bias, and $*$ represents the convolution; $g()$ is for incentive function. The excitation function is introduced into deep neural network, so that it can deal with some complex nonlinear problems better [23]. The commonly used excitation functions are S-shape function, hyperbolic tangent function, linear integral function and so on. Since ReLU can effectively avoid gradient disappearance and gradient explosion compared with existing excitation methods:

$$g(\mu) = \max(0, \mu) \quad (5)$$

Formula (4) is used for the convolution operation of each feature matrix in each order convolution window, then the following new feature graph can be obtained:

$$C = (C_1, C_2, L, C_{13})^T \quad (6)$$

C is a new feature map generated by convolution kernel after convolution operation. Since the eigenmatrix has 13 lines, the convolution core sets the height of 1 and the step of 1, so a feature graph is scanned 13 times from the top to the bottom, that is, the height of the feature graph Z is 13; When the width of the convolution core is set to 1,3, and 5, the width of the feature graph is determined by the size of the convolution core used [24].

## C. Design of fully connected stage and hierarchical stage

The neuron in front of it is:

$$g_m = \sum_i \eta_{mi} x_i + \varepsilon_m \quad (7)$$

$g_m$ is node of the $m$ neuron in the fully connected layer. $i$ nodes above $x_i$. $\eta_{mi}$ is the weight of $mi$. Where $\varepsilon_m$ is the preference for the $m$ neuron [25]. This paper presents a method based on fuzzy cluster analysis [26]. The standardized index function Softmax is used to stimulate the classification level. The probability $Q$ of an input vector $j$ belonging to class $X$ is:

$$Q(g = j \mid X, \eta, \varepsilon) = \frac{e^{\eta_j^* X + \varepsilon_j}}{\sum_{j=1}^{n} e^{\eta_j^* X + \varepsilon_j}} \quad (8)$$

$n$ is the number of categories; $\eta_j$ and $\varepsilon_j$ are weights and biases.

## D. Other Designs

The neural network is divided into two steps: 1) the damage amount is obtained by the forward transfer method; 2) According to the lost value, the recursive method is used to reverse transfer each unit to achieve the optimization of each unit. The loss function is a parameter that measures the difference between actual data and expected data during forward transfer [27]:

$$L(\alpha,\beta) = -(\beta(x)lb\alpha(x)) - \\ (1-\beta(x))lb(1-\alpha(x)) \quad (9)$$

$\beta(x)$ is the true tag associated with input property $x$; $\alpha(x)$ is the probability obtained after the model predicts $x$. When the mutual entropy value of the network is low, it indicates that the actual situation of the network is more consistent with the actual situation of the model, so that the overall learning of the network is closer to the purpose of optimal control [28].

## III. RESULT ANALYSIS

The accuracy of 1D-CNN model is tested by the method of cross test. The method divides all the collected samples into 10 parts to get a correct rate. 10 samples were taken as the test set, 10 consecutive tests were conducted, and the average scores of the 10 tests were taken as the final evaluation criteria [29]. Figure 2,3 shows the relationship between errors and lost values in statistical logs and cancer disease databases as the number of iterations increases.

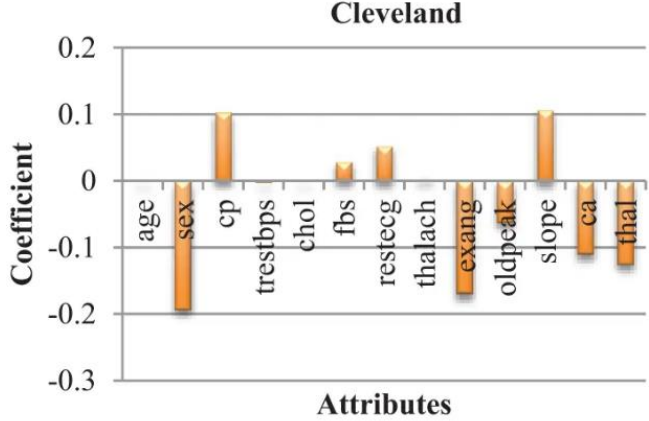
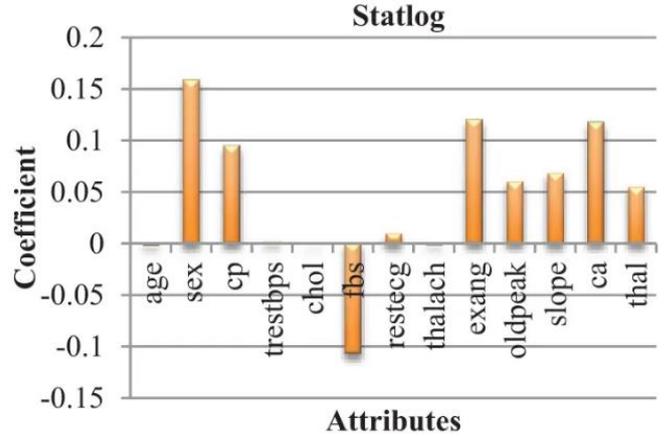

Fig. 2. Statlog dataset model effect

As shown in Figure 2, on the statistical log, its training accuracy can reach 100%, and after more than 30 cycles, its reduction gradually becomes smaller, and the accuracy at this stage is also very stable. Figure 3 shows the effectiveness of the proposed method in the heart disease database [30]. After 33 repetitions, the accuracy rate and loss rate decrease gradually become stable and finally reach below 0.2.

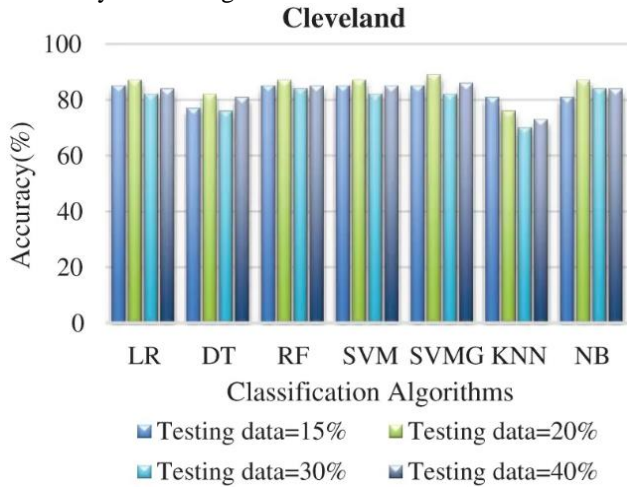
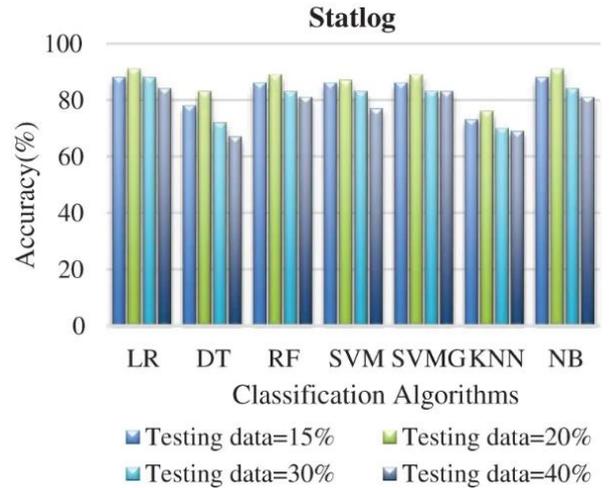

Fig. 3. Model effect of Heart disease database dataset

This project intends to use 1D-CNN model, logistic regression model based on dummy variables, and neural network model based on PSO to conduct independent tests on the above three models to verify their accuracy [31]. The results are shown in Table 2.

TABLE I. COMPARISON OF PREDICTION ACCURACY OF DIFFERENT MODELS

| Model | Accuracy rate /% | |
|---|---|---|
| | Statlog | Heart disease database |
| Dv-Logistic | 85.58 | 85.73 |
| PSO-ELM | 91.99 | 93.38 |
| 1D-CNN | 97.25 | 98.42 |

Compared with Dv-logistic model, the accuracy of PSO-ELM in statistics and heart disease database increased by 6.22% and 7.65%. Compared with PSO-ELM method, the prediction accuracy of 1D-CNN algorithm is 5.25% and 5.05% higher than that of PSO-ELM algorithm.

The main factors that lead to the low accuracy of the existing method are: 1) It is established on the premise of statistics and evidence-based medicine theory, and it has high computational efficiency when applied to known mathematical equations; However, in the case of a large sample dimension, underfitting phenomenon often occurs, which affects the accuracy of classification. 2) Although the PSO method can effectively solve the problems of large error and low accuracy caused by the randomness of the initial value of the trainer, its learning mechanism is limited and its improvement effect is not great because it is still a shallow neural network in nature. The above studies show that the prediction method of 1D-CNN can better overcome the problems of poor effect and low accuracy in the modeling of multi-class and multi-class complex factors, so that it can show more outstanding performance in the prediction of cardiovascular disease.

## IV. CONCLUSION

This project intends to study the multi-parametric cardiovascular disease risk early warning method based on deep neural network, and research the multi-parametric cardiovascular disease risk early warning method, so as to efficiently solve complex problems on the premise of ensuring accuracy. Our methodology demonstrates a compelling case for the utilization of CNNs in handling multi-parametric data, presenting a promising alternative to conventional risk assessment tools. By exploiting the inherent spatial hierarchy in physiological and symptom indicators through convolutional and pooling operations, our model offers an enriched feature representation leading to improved early warning capabilities. Moreover, the optimization through the Adam algorithm and the implementation of a soft classifier refine the model's performance, thus empowering healthcare professionals with a more accurate and non-invasive diagnostic aid. In the broader scope, this research underlines the potential of deep learning to revolutionize predictive healthcare analytics. Future work should focus on expanding the model's interpretability and scalability, integrating multi-source information for a comprehensive understanding of cardiovascular risk factors. Moreover, personalized models considering patient-specific variations and the interplay of treatment conditions could be developed to further individualize risk predictions.